\documentclass[10pt,twocolumn,letterpaper]{article}

\usepackage{iccv}
\usepackage{times}
\usepackage{epsfig}
\usepackage{graphicx}
\usepackage{amsmath}
\usepackage{amssymb}

\usepackage{enumitem}
\usepackage[square,numbers]{natbib}
\usepackage{booktabs}
\usepackage{colortbl}
\usepackage{ifthen}
\usepackage{xcolor}
\usepackage[format=plain,labelformat=simple,labelsep=period,font=small,compatibility=false]{caption}
\usepackage[font=footnotesize,skip=3pt,subrefformat=parens]{subcaption}
\usepackage{multirow}
\usepackage{authblk}

\DeclareMathOperator*{\argmax}{arg\,max}

\usepackage[breaklinks=true,bookmarks=false,hidelinks]{hyperref}

\usepackage[capitalize]{cleveref}

\iccvfinalcopy %

\def\iccvPaperID{12473} %

\ificcvfinal\fi

\newboolean{include-notes}
\setboolean{include-notes}{false}
\newcommand{\suggest}[1]{\ifthenelse{\boolean{include-notes}}{\textcolor{red}{{#1}}}{#1}}
\newcommand{\rami}[1]{\ifthenelse{\boolean{include-notes}}
 {\textcolor{blue}{\textbf{[#1]}}}{}}
\newcommand{\ari}[1]{\ifthenelse{\boolean{include-notes}}
 {\textcolor{cyan}{\textbf{[SA: #1]}}}{}}

\begin{document}

\title{MotionLM: Multi-Agent Motion Forecasting as Language Modeling}

\renewcommand\and{\hspace{1em}}
\author{Ari Seff
\and
Brian Cera
\and
Dian Chen\thanks{Work done during an internship at Waymo.}
\and
Mason Ng
\and
Aurick Zhou
\and
Nigamaa Nayakanti \\
\and
Khaled S. Refaat
\and
Rami Al-Rfou
\and
Benjamin Sapp
}

\affil{\vspace{0.5em}Waymo}

\maketitle

{\let\thefootnote\relax\footnote{{Contact: \texttt{\{aseff, bensapp\}@waymo.com}}}}

\ificcvfinal\thispagestyle{empty}\fi

\begin{abstract}
Reliable forecasting of the future behavior of road agents is a critical component to safe planning in autonomous vehicles.
Here, we represent continuous trajectories as sequences of discrete motion tokens and cast multi-agent motion prediction as a language modeling task over this domain.
Our model, MotionLM, provides several advantages:
First, it does not require anchors or explicit latent variable optimization to learn multimodal distributions.
Instead, we leverage a single standard language modeling objective, maximizing the average log probability over sequence tokens.
Second, our approach bypasses post-hoc interaction heuristics where individual agent trajectory generation is conducted prior to interactive scoring.
Instead, MotionLM produces joint distributions over interactive agent futures in a single autoregressive decoding process.
In addition, the model's sequential factorization enables temporally causal conditional rollouts.
The proposed approach establishes new state-of-the-art performance for multi-agent motion prediction on the Waymo Open Motion Dataset, ranking 1\textsuperscript{st} on the interactive challenge leaderboard.
\end{abstract}

\section{Introduction}

\begin{figure}
\centering
\includegraphics[width=0.95\linewidth]{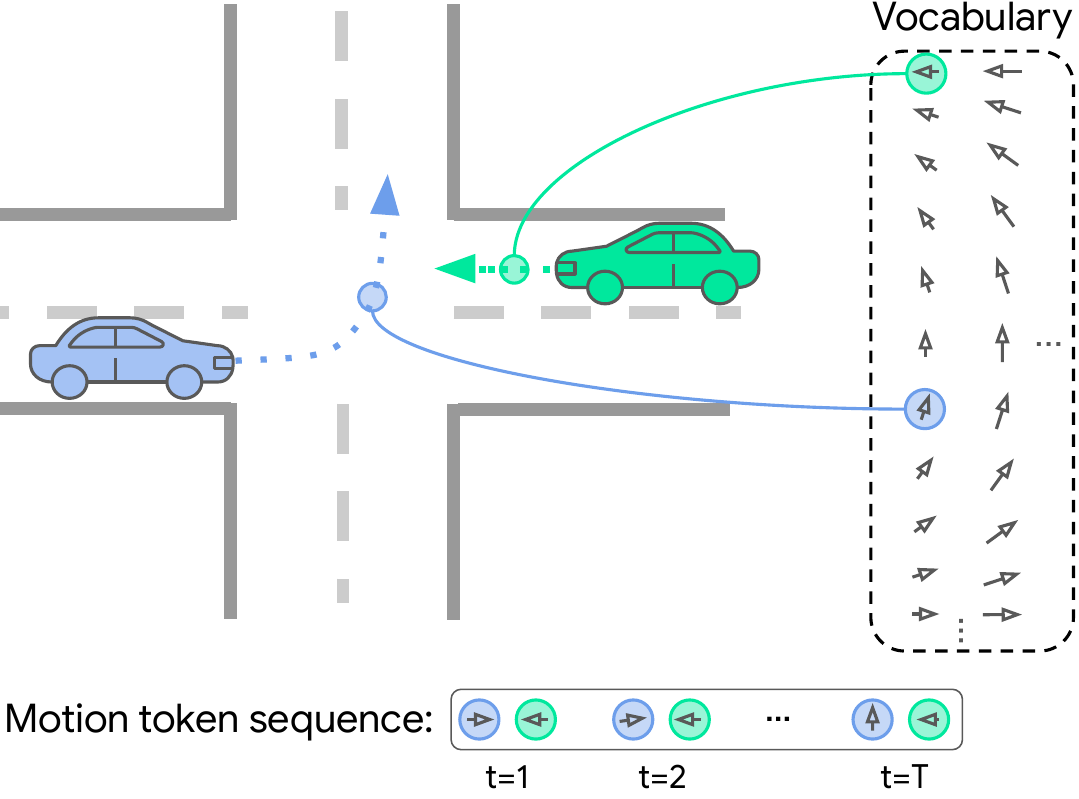}
\caption{Our model autoregressively generates sequences of discrete motion tokens for a set of agents to produce consistent interactive trajectory forecasts.}
\label{fig:teaser}
\end{figure}

Modern sequence models often employ a next-token prediction objective that incorporates minimal domain-specific assumptions.
For example, autoregressive language models \citep{GPT3,PALM} are pre-trained to maximize the probability of the next observed subword conditioned on the previous text; there is no predefined notion of parsing or syntax built in.
This approach has found success in continuous domains as well, such as audio \citep{AudioLM} and image generation \cite{Parti}.
Leveraging the flexibility of arbitrary categorical distributions, the above works represent continuous data with a set of discrete tokens, reminiscent of language model vocabularies.

In driving scenarios, road users may be likened to participants in a constant dialogue, continuously exchanging a dynamic series of actions and reactions mirroring the fluidity %
of communication. 
Navigating this rich web of interactions requires the ability to anticipate the likely maneuvers and responses of the involved actors.
Just as today's language models can capture sophisticated distributions over conversations, can we leverage similar sequence models to forecast the behavior of road agents?

A common simplification to modeling the full future world state has been to decompose the joint distribution of agent behavior into independent per-agent marginal distributions.
Although there has been much progress on this task \citep{multipath,MultiPath++,mtp,lanegcn,wayformer,intentnet,spagnn,WIMP}, marginal predictions are insufficient as inputs to a planning system; they do not represent the future dependencies between the actions of different agents, leading to inconsistent scene-level forecasting.

Of the existing joint prediction approaches, some apply a separation between marginal trajectory generation and interactive scoring \citep{MTR,M2I,JFP}. %
For example, \citet{JFP} initially produce a small set of marginal trajectories for each agent independently, before assigning a learned potential to each inter-agent trajectory pair through a belief propagation algorithm.
\citet{M2I} use a manual heuristic to tag agents as either influencers or reactors, and then pairs marginal and conditional predictions to form joint predictions.

We also note that because these approaches do not explicitly model temporal dependencies within trajectories, their conditional forecasts may be more susceptible to spurious correlations, leading to less realistic reaction predictions.
For example, these models can capture the \emph{correlation} between a lead agent decelerating and a trailing agent decelerating, %
but may fail to infer which one is likely causing the other to slow down.
In contrast, previous joint models employing an autoregressive factorization, e.g., \citep{PRECOG,MFP,trajectron++}, do respect future temporal dependencies.
These models have generally relied on explicit latent variables for diversity, optimized via either an evidence lower bound or normalizing flow.

In this work, we combine trajectory generation and interaction modeling in a single, temporally causal, decoding process over discrete motion tokens (\cref{fig:teaser}), leveraging a simple training objective inspired by autoregressive language models.
Our model, MotionLM, is trained to directly maximize the log probability of these token sequences among interacting agents.
At inference time, joint trajectories are produced step-by-step, where interacting agents sample tokens simultaneously, attend to one another, and repeat.
In contrast to previous approaches which manually enforce trajectory multimodality during training, our model is entirely latent variable and anchor-free, with multimodality emerging solely as a characteristic of sampling.
MotionLM may be applied to several downstream behavior prediction tasks, including marginal, joint, and conditional predictions. %

\begin{figure*}
\centering
\includegraphics[width=0.98\linewidth]{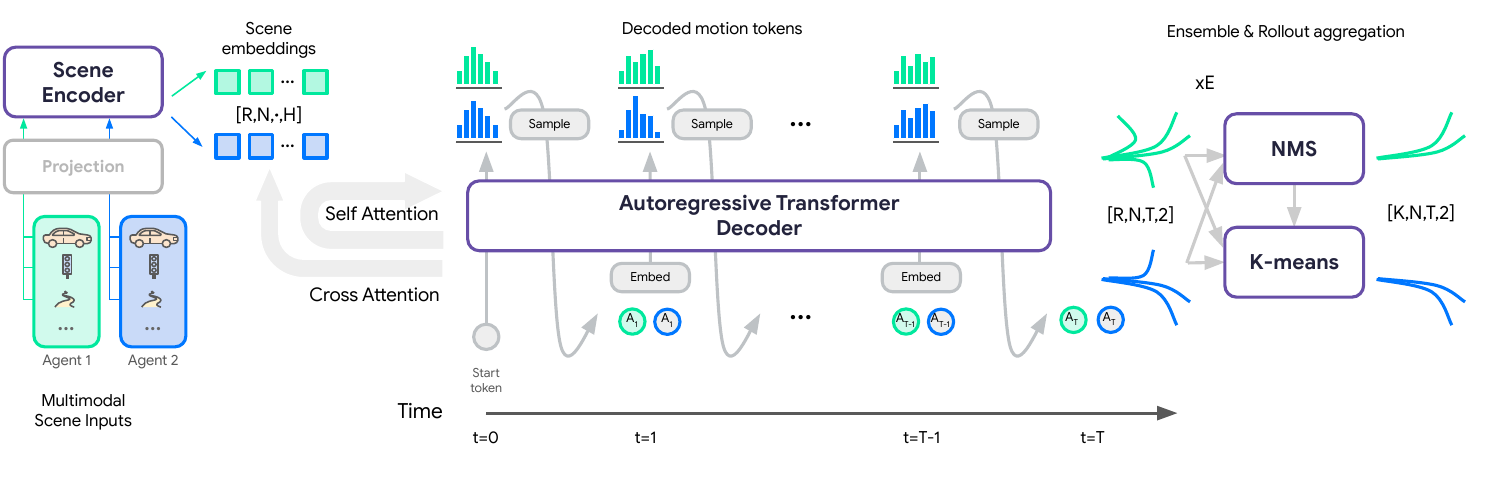}
\caption{MotionLM architecture. We first encode heterogeneous scene features relative to each modeled agent (left) as scene embeddings of shape $R, N, \cdot ,H$. Here, $R$ refers to the number of rollouts, $N$ refers to the number of (jointly modeled) agents, and $H$ is the dimensionality of each embedding. We repeat the embeddings $R$ times in the batch dimension for parallel sampling during inference. Next, a trajectory decoder autoregressively rolls out $T$ discrete motion tokens for multiple agents in a temporally causal manner (center). Finally, representative modes of the rollouts may be recovered via a simple aggregation utilizing k-means clustering initialized with non-maximum suppression (right).}
\label{fig:overview}
\end{figure*}

This work makes the following contributions:
\begin{enumerate}
    \item We cast multi-agent motion forecasting as a language modeling task, introducing a temporally causal decoder over discrete motion tokens trained with a causal language modeling loss.
    \item We pair sampling from our model with a simple rollout aggregation scheme that facilitates weighted mode identification for joint trajectories, establishing new state-of-the-art performance on the Waymo Open Motion Dataset interaction prediction challenge (6\% improvement in the ranking joint mAP metric).
    \item We perform extensive ablations of our approach as well as analysis of its temporally causal conditional predictions, which are largely unsupported by current joint forecasting models.
\end{enumerate}

\section{Related work}

\paragraph{Marginal trajectory prediction.} Behavior predictors are often evaluated on their predictions for individual agents, e.g., in recent motion forecasting benchmarks ~\citep{WOMD,argoverse,nuscenes,interaction-dataset,sdd-dataset}.
Previous methods process the rasterized scene with CNNs \citep{multipath,intentnet,mtp,rules_of_road}; the more recent works represent scenes with points and polygraphs and process them with GNNs~\citep{spagnn,lanegcn,MultiPath++,mpa} or transformers~\citep{wayformer,MTR,hdgt}.
To handle the multimodality of future trajectories, some models manually enforce diversity via predefined anchors \citep{multipath,intentnet} or intention points \citep{MTR,tnt,densetnt}.
Other works learn diverse modes with latent variable modeling, e.g., ~\citep{desire}. 

While these works produce multimodal future trajectories of individual agents, they only capture the marginal distributions of the possible agent futures and do not model the interactions among agents.

\paragraph{Interactive trajectory prediction.} Interactive behavior predictors model the joint distribution of agents' futures.
This task has been far less studied than marginal motion prediction.
For example, the Waymo Open Motion Dataset (WOMD) \citep{WOMD} challenge leaderboard currently has 71 published entries for marginal prediction compared to only 14 for interaction prediction.

\citet{SceneTransformer} models the distribution of future trajectories with a transformer-based mixture model outputting joint modes.
To avoid the exponential blow-up from a full joint model, ~\citet{JFP} models pairwise joint distributions.
\citet{CBP,PiP,M2I} consider conditional predictions by exposing the future trajectory of one agent when predicting for another agent.
\citet{MTR} derives joint probabilities by simply multiplying marginal trajectory probabilities, essentially treating agents as independent, which may limit accuracy.
\citet{lookout,ILVM,AutoBots} reduce the full-fledged joint distribution using global latent variables.
Unlike our autoregressive factorization, the above models typically follow ``one-shot'' (parallel across time) factorizations and do not explicitly model temporally causal interactions.

\paragraph{Autoregressive trajectory prediction.}
Autoregressive behavior predictors generate trajectories at intervals to produce scene-consistent multi-agent trajectories.
\citet{PRECOG,MFP,LatentFormer,trajectron++,AgentFormer} predict multi-agent future trajectories using latent variable models.
\citet{KEMP} explores autoregressively outputting keyframes via mixtures of Gaussians prior to filling in the remaining states.
In \citep{Symphony}, an adversarial objective is combined with parallel beam search to learn multi-agent rollouts.
Unlike most autoregressive trajectory predictors, our method does not rely on latent variables or beam search and generates multimodal joint trajectories by directly sampling from a learned distribution of discrete motion token sequences.

\paragraph{Discrete sequence modeling in continuous domains.}
When generating sequences in continuous domains, one effective approach is to discretize the output space and predict categorical distributions at each step.

For example, in image generation, \citet{PixelRNN} sequentially predict the uniformly discretized pixel values for each channel and found this to perform better than outputting continuous values directly.
Multiple works on generating images from text such as \citep{DALL_E} and \citep{Parti} use a two-stage process with a learned tokenizer to map images to discrete tokens and an autoregressive model to predict the discrete tokens given the text prompt.
For audio generation, WaveNet \citep{WaveNet} applies a $\mu$-law transformation before discretizing.
\citet{AudioLM} learn a hierarchical tokenizer/detokenizer, with the main transformer sequence model operating on the intermediate discrete tokens.
When generating polygonal meshes, \citet{PolyGen} uniformly quantize the coordinates of each vertex.
In MotionLM, we employ a simple uniform quantization of axis-aligned deltas between consecutive waypoints of agent trajectories.

\section{MotionLM} \label{sec:method}

We aim to model a distribution over multi-agent interactions in a general manner that can be applied to distinct downstream tasks, including marginal, joint, and conditional forecasting.
This requires an expressive generative framework capable of capturing the substantial multimodality in driving scenarios.
In addition, we take consideration here to preserve temporal dependencies; i.e., inference in our model follows a directed acyclic graph with the parents of every node residing earlier in time and children residing later (\cref{sec:temporal_causality}, \cref{fig:temporal_causal_structure}).
This enables conditional forecasts that more closely resemble causal interventions \citep{Pearl09} by eliminating certain spurious correlations that can otherwise result from disobeying temporal causality\footnote{We make no claims that our model is capable of directly modeling causal relationships (due to the theoretical limits of purely observational data and unobserved confounders). Here, we solely take care to avoid breaking temporal causality.}.
We observe that joint models that do not preserve temporal dependencies may have a limited ability to predict realistic agent reactions -- a key use in planning (\cref{sec:conditional_rollouts_exp}).
To this end, we leverage an autoregressive factorization of our future decoder, where agents' motion tokens are conditionally dependent on all previously sampled tokens and trajectories are rolled out sequentially (\cref{fig:overview}).

Let $S$ represent the input data for a given scenario.
This may include context such as roadgraph elements, traffic light states, as well as features describing road agents (e.g., vehicles, cyclists, and pedestrians) and their recent histories, all provided at the current timestep $t=0$.
Our task is to generate predictions for joint agent states $Y_t \doteq \{y^1_t, y^2_t, ..., y^N_t\}$ for $N$ agents of interest at future timesteps $t=1, ..., T$.
Rather than complete states, these future state targets are typically two-dimensional waypoints (i.e., $(x,y)$ coordinates), with $T$ waypoints forming the full ground truth trajectory for an individual agent.

\subsection{Joint probabilistic rollouts}
\begin{figure*}
\centering
\includegraphics[width=0.95\linewidth]{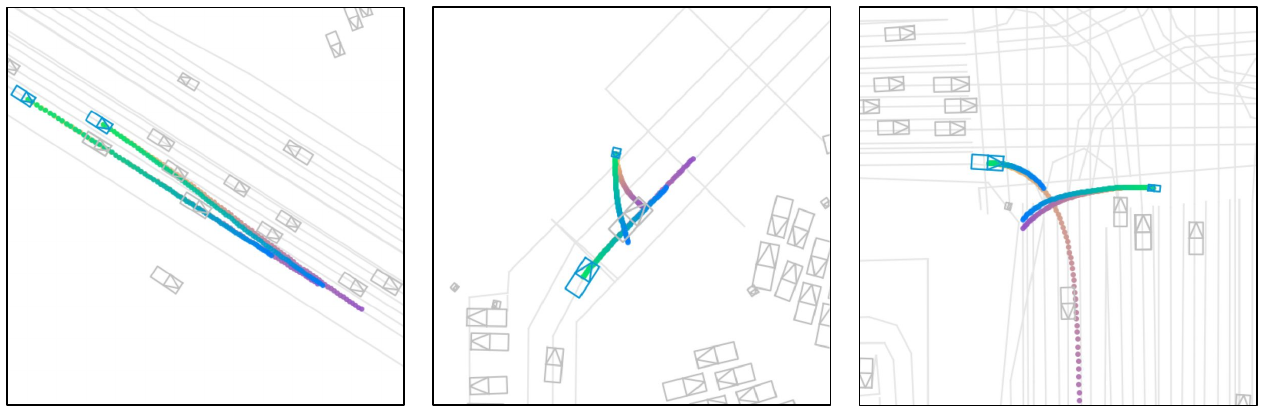}
\caption{Displayed are the top two predicted joint rollout modes for three WOMD scenes. Color gradients indicate time progression from $t=0{\rm s}$ to $t=8{\rm s}$, with the greatest probability joint mode transitioning from green to blue and the secondary joint mode transitioning from orange to purple. Three types of interactions are observed: an agent in the adjacent lane yields to the lane-changing agent according to the timing of the lane change (left), a pedestrian walks behind the passing vehicle according to the progress of the vehicle (center), the turning vehicle either yields to the crossing cyclist (most probable mode) or turns before the cyclist approaches (secondary mode) (right). }
\label{fig:top_2_joint_viz}
\end{figure*}

In our modeling framework, we sample a predicted action for each target agent at each future timestep.
These actions are formulated as discrete motion tokens from a finite vocabulary, as described later in \cref{sec:traj_decoder}.
Let $a^n_t$ represent the target action (derived from the ground truth waypoints) for the $n$th agent at time $t$, with $A_t \doteq \{a^1_t, a^2_t, ..., a^N_t\}$ representing the set of target actions for all agents at time $t$.

\paragraph{Factorization.}
We factorize the distribution over joint future action sequences as a product of conditionals:
\begin{equation}
    p_\theta(A_1, A_2, ... A_T \mid S) = \prod_{t=1}^T p_\theta(A_t \mid A_{<t}, S),
\end{equation}
\begin{equation} \label{eq:conditional_indep}
    p_\theta(A_t \mid A_{<t}, S) = \prod_{n=1}^N p_\theta(a^n_t \mid A_{<t}, S).
\end{equation}
Similar to \citep{PRECOG,MFP}, \cref{eq:conditional_indep} represents the fact that we treat agent actions as conditionally independent at time $t$, given the previous actions and scene context.
This aligns empirically with real-world driving over short time intervals; e.g., non-impaired human drivers generally require at least 500 ms to release the accelerator in response to a vehicle braking ahead (\citep{ResponseTiming}).  %
In our experiments, we find 2 Hz reactions to be sufficient to surpass state-of-the-art joint prediction models.

We note that our model's factorization is entirely latent variable free; multimodal predictions stem purely from categorical token sampling at each rollout timestep.

\paragraph{Training objective.}
MotionLM is formulated as a generative model trained to match the joint distribution of observed agent behavior.
Specifically, we follow a maximum likelihoood objective over multi-agent action sequences:
\begin{equation}
    \argmax_\theta \prod_{t=1}^T p_\theta(A_t \mid A_{<t}, S)
\end{equation}

Similar to the typical training setup of modern language models, we utilize ``teacher-forcing'' where previous ground truth (not predicted) tokens are provided at each timestep, which tends to increase stability and avoids sampling during training.
We note that this applies to all target agents; in training, each target agent is exposed to ground truth action sequence prefixes for all target agents prior to the current timestep.
This naturally allows for temporal parallelization when using modern attention-based architectures such as transformers \citep{transformers}.

Our model is subject to the same theoretical limitations as general imitation learning frameworks (e.g., compounding error \citep{DAGGER} and self-delusions due to unobserved confounders \citep{ortegadelusions2021}).
However, we find that, in practice, these do not prevent strong performance on forecasting tasks.

\subsection{Model implementation}
Our model consists of two main networks, an encoder which processes initial scene elements followed by a trajectory decoder which performs both cross-attention to the scene encodings and self-attention along agent motion tokens, following a transformer architecture \citep{transformers}.

\subsubsection{Scene encoder} \label{sec:scene_encoder}
The scene encoder (\cref{fig:overview}, left) is tasked with processing information from several input modalities, including the roadgraph, traffic light states, and history of surrounding agents' trajectories.
Here, we follow the design of the early fusion network proposed by \citep{wayformer} as the scene encoding backbone of our model. Early fusion is particularly chosen because of its flexibility to process all modalities together with minimal inductive bias.

The features above are extracted with respect to each modeled agent's frame of reference.
Input tensors are then fed to a stack of self-attention layers that exchange information across all past timesteps and agents.
In the first layer, latent queries cross-attend to the original inputs in order to reduce the set of vectors being processed to a manageable number, similar to \citep{settransformer,perceiver}.
For additional details, see \citep{wayformer}.

\subsubsection{Joint trajectory decoder}  \label{sec:traj_decoder}
Our trajectory decoder (\cref{fig:overview}, center) is tasked with generating sequences of motion tokens for multiple agents.

\paragraph{Discrete motion tokens.}
We elect to transform trajectories comprised of continuous waypoints into sequences of discrete tokens.
This enables treating sampling purely as a classification task at each timestep, implemented via a standard softmax layer.
Discretizing continuous targets in this manner has proven effective in other inherently continuous domains, e.g., in audio generation \citep{WaveNet} and mesh generation \citep{PolyGen}.  
We suspect that discrete motion tokens also naturally hide some precision from the model, possibly mitigating compounding error effects that could arise from imperfect continuous value prediction.
Likewise, we did not find it necessary to manually add any noise to the ground truth teacher-forced trajectories (e.g., as is done in \citep{DeepStruct}).

\paragraph{Quantization.}
To extract target discrete tokens, we begin by normalizing each agent's ground truth trajectory with respect to the position and heading of the agent at time $t=0$ of the scenario. %
We then parameterize a uniformly quantized $(\Delta x, \Delta y)$ vocabulary according to a total number of per-coordinate bins as well as maximum and minimum delta values.
A continuous, single-coordinate delta action can then be mapped to a corresponding index $\in [0, {\rm num\_bins} - 1]$, resulting in two indices for a complete $(\Delta x, \Delta y)$ action per step.
In order to extract actions that accurately reconstruct an entire trajectory, we employ a greedy search, sequentially selecting the quantized actions that reconstruct the next waypoint coordinates with minimum error.

We wrap the delta actions with a ``Verlet'' step where a zero action indicates that the same delta index should be used as the previous step (as \citep{PRECOG} does for continuous states).
As agent velocities tend to change smoothly between consecutive timesteps, this helps reduce the total vocabulary size, simplifying the dynamics of training.
Finally, to maintain only $T$ sequential predictions, we collapse the per-coordinate actions to a single integer indexing into their Cartesian product. %
In practice, for the models presented here, we use 13 tokens per coordinate with $13^2=169$ total discrete tokens available in the vocabulary (see \cref{sec:action_params} for further details).

We compute a learned value embedding and two learned positional embeddings (representing the timestep and agent identity) for each discrete motion token, which are combined via an element-wise sum prior to being input to the transformer decoder.

\paragraph{Flattened agent-time self-attention.} 
We elect to include a single self-attention mechanism in the decoder that operates along flattened sequences of all modeled agents' motion tokens over time.
So, given a target sequence of length $T$ for each of $N$ agents, we perform self-attention over $NT$ elements.
While this does mean that these self-attended sequences grow linearly in the number of jointly modeled agents, we note that the absolute sequence length here is still quite small (length 32 for the WOMD interactive split -- 8 sec. prediction at 2 Hz for 2 agents). %
Separate passes of factorized agent and time attention are also possible \citep{SceneTransformer}, but we use a single pass here for simplicity.

\paragraph{Ego agent reference frames.}
To facilitate cross-attention to the agent-centric feature encodings (\cref{sec:scene_encoder}), we represent the flattened token sequence once for each modeled agent.
Each modeled agent is treated as the ``ego'' agent once, and cross-attention is performed on that agent's scene features.
Collapsing the ego agents into the batch dimension allows parallelization during training and inference.

\subsection{Enforcing temporal causality}  \label{sec:temporal_causality}
Our autoregressive factorization naturally respects temporal dependencies during joint rollouts; motion token sampling for any particular agent is affected only by past tokens (from any agent) and unaffected by future ones.
When training, we require a mask to ensure that the self-attention operation only updates representations at each step accordingly.
As shown in \cref{fig:causal_attention} (appendix), this attention mask exhibits a blocked, staircase pattern, exposing all agents to each other's histories only up to the preceding step.

\begin{figure*}
\begin{subfigure}[b]{0.33\textwidth}
\raggedright
\includegraphics[width=0.92\textwidth,page=1]{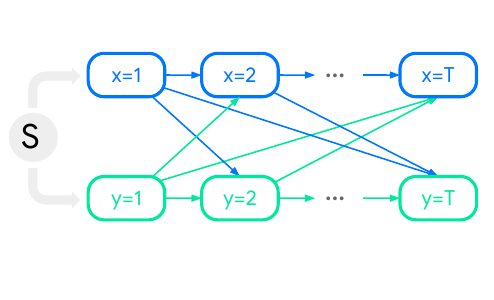}
\caption{Causal Bayesian Network for joint rollouts.}
\end{subfigure}%
\begin{subfigure}[b]{0.33\textwidth}
\includegraphics[width=0.92\textwidth,page=2]{figs/causal_graph5.pdf}
\caption{Post-Intervention Causal Bayesian Network.}
\end{subfigure}%
\begin{subfigure}[b]{0.33\textwidth}
\raggedleft
\includegraphics[width=0.92\textwidth,page=3]{figs/causal_graph5.pdf}
\caption{Acausal conditioning.}
\end{subfigure}%
\caption{A Causal Bayesian network representation for joint rollouts (left), post-intervention Causal Bayesian network (center), and acausal conditioning (right). 
Solid lines indicate temporally causal dependencies while dashed lines indicate acausal information flow.
Models without temporal dependency constraints will support acausal conditioning but not temporally causal conditioning, which can be problematic when attempting to predict agent reactions.}
\label{fig:temporal_causal_structure}
\end{figure*}

\paragraph{Temporally causal conditioning.}
As described earlier, a particular benefit of this factorization is the ability to query for temporally causal conditional rollouts (\cref{fig:temporal_causal_structure}).
In this setting, we fix a query agent to take some sequence of actions and only roll out the other agents.

We may view this as an approximation of computing causal interventions \citep{Pearl09} in the absence of confounders; interventions cannot be learned purely through observational data in general (due to the possible presence of unobserved confounders), but our model's factorization at least eliminates certain spurious correlations arising from breaking temporal causality.

In~\cref{fig:temporal_causal_structure} (a), we show an example of a Causal Bayesian network governing joint rollouts. Applying an intervention to nodes $x=1,...T$, by deleting their incoming edges, results in a post-intervention Bayesian network depicted in~\cref{fig:temporal_causal_structure} (b), which obeys temporal causality. On the other hand, acausal conditioning (\cref{fig:temporal_causal_structure} (c)) results in non-causal information flow, where node $x=i$ affects our belief about node $y=j$ for $i \ge j$.

\subsection{Rollout aggregation}
\label{sec:rollout_aggregation}
Joint motion forecasting benchmark tasks like WOMD~\cite{WOMD} require a compact representation of the joint future distribution in the form of a small number of joint ``modes''.
Each mode is assigned a probability and might correspond to a specific homotopic outcome (e.g., pass/yield) or more subtle differences in speed/geometry.
Here, we aggregate rollouts to achieve two primary goals: 1) uncover the underlying modes of the distribution and 2) estimate the probability of each mode.
Specifically, we follow the non-maximum suppression (NMS) aggregation scheme described in \citep{MultiPath++}, but extend it to the joint setting by ensuring that all agent predictions reside within a given distance threshold to the corresponding cluster.
In addition, we leverage model ensembling to account for epistemic uncertainty and further improve the quality of the predictions, combining rollouts from independently trained replicas prior to the aggregation step.

\section{Experiments} 
We evaluate MotionLM on marginal and joint motion forecasting benchmarks, examine its conditional predictions and conduct ablations of our modeling choices.

\subsection{Datasets}

\paragraph{Waymo Open Motion Dataset (WOMD).}
WOMD \citep{WOMD} is a collection of 103k 20-second scenarios collected from real-world driving in urban and suburban environments. 
Segments are divided into 1.1M examples consisting of 9-second windows of interest, where the first second serves as input context and the remaining 8-seconds are the prediction target.
Map features such as traffic signal states and lane features are provided along with agent states such as position, velocity, acceleration, and bounding boxes.

\paragraph{Marginal and interactive prediction challenges.}
For the marginal prediction challenge, six trajectories must be output by the model for each target agent, along with likelihoods of each mode.
For the interactive challenge, two interacting agents are labeled in each test example.
In this case, the model must output six weighted \emph{joint} trajectories.

\subsection{Metrics}

The primary evaluation metrics for the marginal and interactive prediction challenges are soft mAP and mAP, respectively, with miss rate as the secondary metric.
Distance metrics minADE and minFDE provide additional signal on prediction quality.
For the interactive prediction challenge, these metrics refer to scene-level joint calculations.
We also use a custom \emph{prediction overlap} metric (similar to \citep{JFP}) to assess scene-level consistency for joint models.
See \cref{sec:metrics_desciptions} for details on these metrics.

\subsection{Model configuration}
We experiment with up to 8 model replicas and 512 rollouts per replica, assessing performance at various configurations.
For complete action space and model hyperparameter details, see \cref{sec:action_params,sec:model_arch_details}.

\subsection{Quantitative results}

\begin{table*}\small\centering
\scalebox{0.95}{
\begin{tabular}{l|r|r>{\columncolor[gray]{0.85}}r|>{\columncolor[gray]{0.85}}r} \toprule 
 Model                   & minADE ($\downarrow$) & minFDE ($\downarrow$) & Miss Rate ($\downarrow$) & Soft mAP ($\uparrow$) \\ \midrule 
 HDGT \citep{hdgt}                         & 0.7676 & 1.1077 & 0.1325 & 0.3709 \\
 MPA \citep{mpa}                           & 0.5913 & 1.2507 & 0.1603 & 0.3930 \\
 MTR    \citep{MTR}                        & 0.6050 & 1.2207 & 0.1351 & 0.4216 \\
 Wayformer factorized \citep{wayformer}    & \textbf{0.5447} & 1.1255 & 0.1229 & 0.4260 \\
 Wayformer multi-axis \citep{wayformer}    & 0.5454 & 1.1280 & 0.1228 & 0.4335 \\
 MTR-A  \citep{MTR}                        & 0.5640 & 1.1344 & 0.1160 & \textbf{0.4594} \\
 MotionLM (Ours)                     & 0.5509 & \textbf{1.1199} & \textbf{0.1058} & 0.4507 \\
 \bottomrule
\end{tabular}
}
\caption{Marginal prediction performance on WOMD test set. We display metrics averaged over time steps (3, 5, and 8 seconds) and agent types (vehicles, pedestrians, and cyclists). Greyed columns indicate the official ranking metrics for the marginal prediction challenge.}
\label{table:marginal_results}
\end{table*}

\begin{table*}\small\centering
\scalebox{0.95}{
\begin{tabular}{l|r|r>{\columncolor[gray]{0.85}}r|>{\columncolor[gray]{0.85}}r}
\toprule 
 Model                  & minADE ($\downarrow$) & minFDE ($\downarrow$) & Miss Rate ($\downarrow$) & mAP ($\uparrow$) \\ \midrule 
 SceneTransformer (J) \citep{SceneTransformer} & 0.9774 & 2.1892 & 0.4942 & 0.1192 \\
 M2I \citep{M2I} & 1.3506 & 2.8325 & 0.5538 & 0.1239 \\
 DenseTNT \citep{densetnt} & 1.1417 & 2.4904 & 0.5350 & 0.1647 \\
 MTR \cite{MTR}      & 0.9181 & 2.0633 & 0.4411 & 0.2037 \\
 JFP \citep{JFP}     & \textbf{0.8817} & \textbf{1.9905} & 0.4233 & 0.2050 \\
 MotionLM (Ours)                   & 0.8911 & 2.0067 & \textbf{0.4115} & \textbf{0.2178} \\ 
 \bottomrule
\end{tabular}
}
\caption{Joint prediction performance on WOMD interactive test set. We display \emph{scene-level} joint metrics averaged over time steps (3, 5, and 8 seconds) and agent types (vehicles, pedestrians, and cyclists). Greyed columns indicate the official ranking metrics for the challenge.}
\label{table:joint_results}
\end{table*}

\paragraph{Marginal motion prediction.}
As shown in \cref{table:marginal_results}, our model is competitive with the state-of-the-art on WOMD marginal motion prediction (independent per agent).
For the main ranking metric of soft mAP, our model ranks second, less than 2\% behind the score achieved by MTRA \citep{MTR}.
In addition, our model attains a substantially improved miss rate over all prior works, with a relative 9\% reduction compared to the previous state-of-the-art.
The autoregressive rollouts are able to adequately capture the diversity of multimodal future behavior without reliance on trajectory anchors \citep{multipath} or static intention points \citep{MTR}.

\paragraph{Interactive motion prediction.}
Our model achieves state-of-the-art results for the interactive prediction challenge on WOMD, attaining a 6\% relative improvement in mAP and 3\% relative improvement in miss rate (the two official ranking metrics) over the previous top scoring entry, JFP \citep{JFP} (see \cref{table:joint_results}).
In contrast to JFP, our approach does not score pairs of previously constructed marginal trajectories.
but generates joint rollouts directly. %
\cref{fig:top_2_joint_viz} displays example interactions predicted by our model.

\begin{table}\small\centering
\scalebox{0.95}{
\begin{tabular}{c|l|r} \toprule 
 & Model  & Prediction Overlap  ($\downarrow$) \\ \midrule 
 \multirow{4}{2em}{Test} & LSTM Baseline \citep{WOMD}    &  0.07462 \\
 & Scene Transformer \citep{SceneTransformer}    & 0.04336  \\
 & JFP  \citep{JFP}                        & 0.02671 \\
 & MotionLM (joint)                     & \textbf{0.02607} \\ \midrule
 \multirow{2}{2em}{Val} & MotionLM (marginal) & 0.0404 \\
 & MotionLM (joint)  & \textbf{0.0292} \\
 \bottomrule
\end{tabular}
}
\caption{Prediction overlap rates. Displayed is the custom prediction overlap metric for various model configurations on the WOMD interactive test and validation sets.}
\label{table:prediction_overlap}
\end{table}

\cref{table:prediction_overlap} displays prediction overlap rates  for various models on the WOMD interactive test and validation sets (see metric details in \cref{sec:prediction_overlap_description}).
We obtain test set predictions from the authors of \citep{WOMD,SceneTransformer,JFP}.
MotionLM obtains the lowest prediction overlap rate, an indication of scene-consistent predictions.
In addition, on the validation set we evaluate two versions of our model: marginal and joint.
The marginal version does not perform attention across the modeled agents during both training and inference rollouts, while the joint version performs 2 Hz interactive attention.
We see that the marginal version obtains a relative 38\% higher overlap rate than the joint version.
The interactive attention in the joint model allows the agents to more appropriately react to one another.

\subsection{Ablation studies}

\paragraph{Interactive attention frequency.}
To assess the importance of inter-agent reactivity during the joint rollouts, we vary the frequency of the interactive attention operation while keeping other architecture details constant.
For our leaderboard results, we utilize the greatest frequency studied here, 2 Hz.
At the low end of the spectrum, 0.125 Hz corresponds to the agents only observing each other's initial states, and then proceeding with the entire 8-second rollout without communicating again (i.e., marginal rollouts).

\begin{figure}
    \centering
    \includegraphics[width=1.0\linewidth]{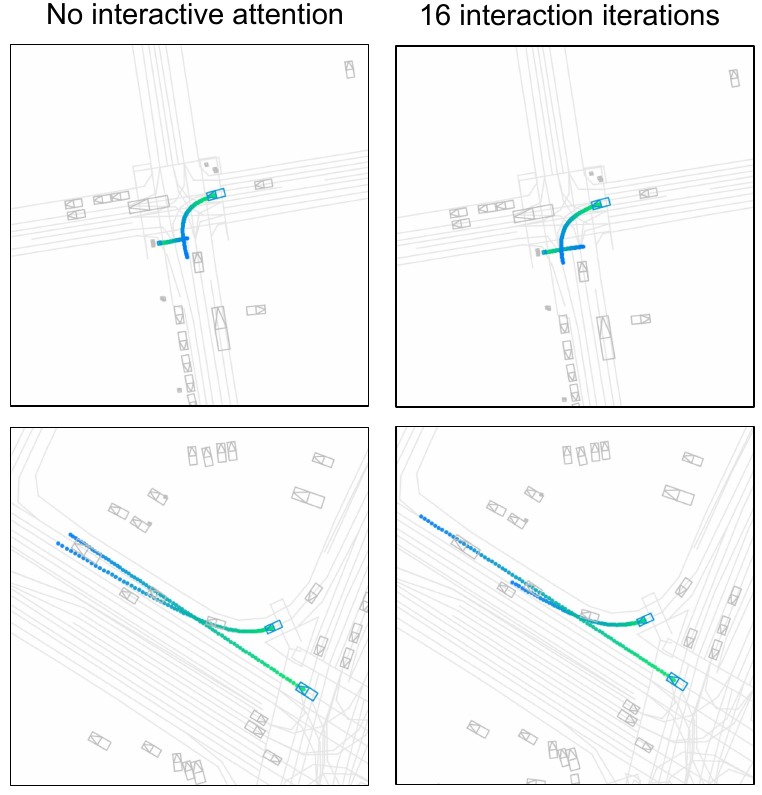}
    \caption{Visualization of the top joint rollout mode at the two extremes of the interactive attention frequencies studied here. With no interactive attention (left), the two modeled agents only attend to each other \emph{once} at the beginning of the 8-second rollout and never again, in contrast to 16 total times for 2 Hz attention (right). The independent rollouts resulting from zero interactive attention can result in scene-inconsistent overlap; e.g., a turning vehicle fails to accommodate a crossing pedestrian (top left) or yield appropriately to a crossing vehicle (bottom left).}
    \label{fig:interact_att_freq_viz}
\end{figure}

Performance metrics generally improve as agents are permitted to interact more frequently (\cref{fig:attention_and_rollouts_ablations} top, \cref{table:interact_att_ablation} in appendix).
Greater interactive attention frequencies not only lead to more accurate joint predictions, but also reduce implausible overlaps (i.e., collisions) between different agents' predictions.
\cref{fig:interact_att_freq_viz} displays examples where the marginal predictions lead to implausible overlap between agents while the joint predictions lead to appropriately separated trajectories.
See supplementary for animated visualizations.

\begin{figure}
\centering
\begin{subfigure}{0.47\textwidth}
\includegraphics[width=\textwidth]{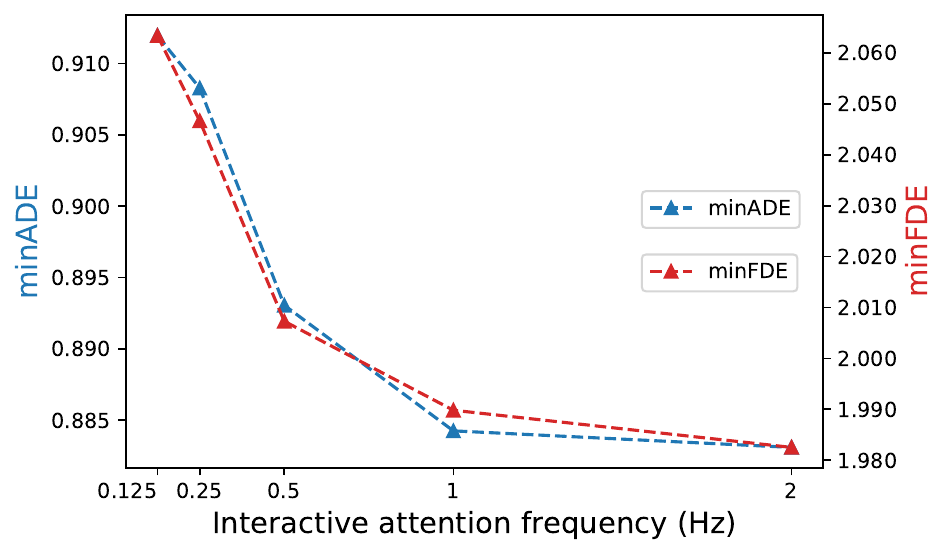}
\end{subfigure}%
\vspace{0.9em}
\begin{subfigure}{0.47\textwidth}
\includegraphics[width=\textwidth]{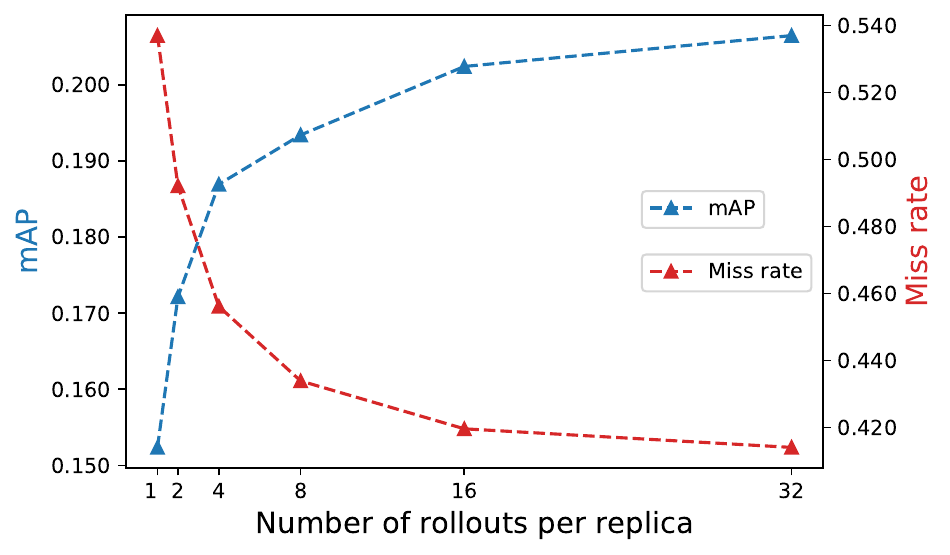}
\end{subfigure}%
\caption{Joint prediction performance across varying interactive attention frequencies (top) and numbers of rollouts per replica (bottom) on the WOMD interactive validation set. Vertical axes display joint (scene-level) metrics for an 8-replica ensemble. See \cref{table:interact_att_ablation,table:num_rollouts_ablation} in the appendix for full parameter ranges and metrics.}
\label{fig:attention_and_rollouts_ablations}
\end{figure}

\paragraph{Number of rollouts.}
Our rollout aggregation requires that we generate a sufficient number of samples from the model in order to faithfully represent the multimodal future distribution.
For this ablation, we vary the number of rollouts generated, but always cluster down to $k=6$ modes for evaluation.
In general, we see performance metrics improve as additional rollouts are utilized (\cref{fig:attention_and_rollouts_ablations}, bottom and \cref{table:num_rollouts_ablation} in appendix).
For our final leaderboard results, we use 512 rollouts per replica, although 32 rollouts is sufficient to surpass the previous top entry on joint mAP.

\begin{table*}[ht!] \small\centering
\scalebox{0.95}{
\begin{tabular}{l|r|r|r|r} \toprule 
 Prediction setting                   & minADE ($\downarrow$) & minFDE ($\downarrow$) & Miss Rate ($\downarrow$) & Soft mAP ($\uparrow$) \\ \midrule 
 Marginal                            & 0.6069 & 1.2236 & 0.1406 & 0.3951 \\
 Temporally causal conditional       & 0.5997 & 1.2034 & 0.1377 & 0.4096 \\
 Acausal conditional                 & 0.5899 & 1.1804 & 0.1338 & 0.4274 \\ \bottomrule
\end{tabular}
}
\caption{Conditional prediction performance. Displayed are marginal (single-agent) metrics across three prediction settings for our model on the WOMD interactive validation set: marginal, temporally causal conditional, and acausal conditional.}
\label{table:conditional_results}
\end{table*}

\subsection{Conditional rollouts} \label{sec:conditional_rollouts_exp}

As described in \cref{sec:temporal_causality}, our model naturally supports ``temporally causal'' conditioning, similarly to previous autoregressive efforts such as \citep{PRECOG,MFP}.
In this setting, we fix one query agent to follow a specified trajectory and stochastically roll out the target agent.
However, we can also modify the model to leak the query agent's full trajectory, acausally exposing its future to the target agent during conditioning.
This resembles the approach to conditional prediction in, e.g., \citep{CBP}, where this acausal conditional distribution is modeled directly, or \citep{JFP}, where this distribution is accessed via inference in an energy-based model.

\begin{figure}
    \centering
    \includegraphics[width=\linewidth]{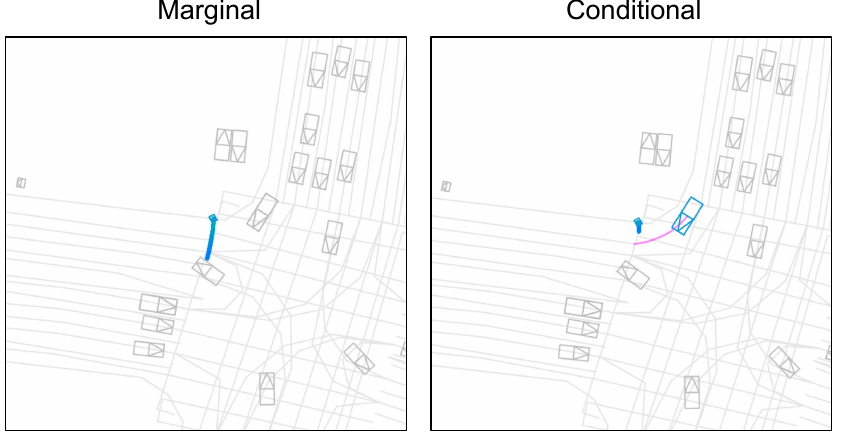}
    \caption{Visualization of the most likely predicted future for the pedestrian in the marginal setting (left) and temporally causal conditional setting (right). When considering the pedestrian independently, the model assigns greatest probability to a trajectory which crosses the road. When conditioned on the the vehicle's ground truth turn (magenta), the pedestrian is instead predicted to yield.}
    \label{fig:conditional_example_viz}
\end{figure}

Here, we assess predictions from our model across three settings: marginal, temporally causal conditional, and acausal conditional (\cref{fig:temporal_causal_structure}).
Quantitatively, we observe that both types of conditioning lead to more accurate predictions for the target agent (\cref{table:conditional_results}, \cref{fig:conditional_example_viz}).
Additionally, we see that acausal conditioning leads to greater improvement than temporally causal conditioning relative to marginal predictions across all metrics, e.g., 8.2\% increase in soft mAP for acausal vs. 3.7\% increase for temporally causal.

Intuitively, the greater improvement for acausal conditioning makes sense as it exposes more information to the model.
However, the better quantitative scores are largely due to predictions that would be deemed nonsensical if interpreted as predicted \emph{reactions} to the query agent.

This can be illustrated in examples where one agent is following another, where typically the lead agent's behavior is causing the trailing agent's reaction, and not vice versa, but this directionality would not be captured with acausal conditoning.
This temporally causal modeling is especially important when utilizing the conditional predictions to evaluate safety for an autonomous vehicle's proposed plans. 
In a scenario where an autonomous vehicle (AV) is stopped behind another agent, planning to move forward into the other agent's current position could be viewed as a safe maneuver with acausal conditioning, as the other agent also moving forward is correlated with (but not caused by) the AV proceeding. 
However, it is typically the lead agent moving forward that causes the trailing AV to proceed, and the AV moving forward on its own would simply rear-end the lead agent.

In the supplementary, we compare examples of predictions in various scenarios for the causal and acausal conditioning schemes.
Models that ignore temporal dependencies during conditioning (e.g., \citep{CBP, JFP}) may succumb to the same incorrect reasoning that the acausal version of our model does.

\section{Conclusion and future work}
In this work, we introduced a method for interactive motion forecasting leveraging multi-agent rollouts over discrete motion tokens, capturing the joint distribution over multimodal futures.
The proposed model establishes new state-of-the-art performance on the WOMD interactive prediction challenge.

Avenues for future work include leveraging the trained model in model-based planning frameworks, allowing a search tree to be formed over the multi-agent action rollouts, or learning amortized value functions from large datasets of scene rollouts.
In addition, we plan to explore distillation strategies from large autoregressive teachers, enabling faster student models to be deployed in latency-critical settings.

\vspace{3em}
\paragraph{Acknowledgements.}
We would like to thank David Weiss, Paul Covington, Ashish Venugopal, Piotr Fidkowski, and Minfa Wang for discussions on autoregressive behavior predictors; Cole Gulino and Brandyn White for advising on interactive modeling; Cheol Park, Wenjie Luo, and Scott Ettinger for assistance with evaluation; Drago Anguelov, Kyriacos Shiarlis, and anonymous reviewers for helpful feedback on the manuscript.

\clearpage
{\small
\bibliographystyle{plainnat}  %
\bibliography{references}
}

\clearpage
\appendix

\section{Motion token vocabulary}  \label{sec:action_params}

\paragraph{Delta action space.}
The models presented in this paper use the following parameters for the discretized delta action space:
\begin{itemize}
    \item Step frequency: 2 Hz
    \item Delta interval (per step): [-18.0 m, 18.0 m]
    \item Number of bins: 128
\end{itemize}
At 2 Hz prediction, a maximum delta magnitude of 18 m covers axis-aligned speeds up to 36 m/s ($\sim$80 mph), $> 99\%$ of the WOMD dataset.

\paragraph{Verlet-wrapped action space.}
Once the above delta action space has the Verlet wrapper applied, we only require 13 bins for each coordinate.
This results in a total of $13^2=169$ total discrete motion tokens that the model can select from the Cartesian product comprising the final vocabulary. 

\paragraph{Sequence lengths.}
For 8-second futures, the model outputs 16 motion tokens for each agent (note that WOMD evaluates predictions at 2 Hz).
For the two-agent interactive split, our flattened agent-time token sequences (\cref{sec:traj_decoder}) have length $2 \times 16 = 32$.

\section{Implementation details}  \label{sec:model_arch_details}
\subsection{Scene encoder}
We follow the design of the early fusion network proposed  by  \cite{wayformer}  as  the  scene  encoding  backbone  of  our model.
The following hyperparameters are used:
\begin{itemize}
    \item Number of layers: 4
    \item Hidden size: 256
    \item Feed-forward network intermediate size: 1024
    \item Number of attention heads: 4
    \item Number of latent queries: 92
    \item Activation: ReLU
\end{itemize}

\subsection{Trajectory decoder}
To autoregressively decode motion token sequences, we utilize a causal transformer decoder that takes in the motion tokens as queries, and the scene encodings as context. 
We use the following model hyperparameters:
\begin{itemize}
    \item Number of layers: 4
    \item Hidden size: 256
    \item Feed-forward network intermediate size: 1024
    \item Number of attention heads: 4
    \item Activation: ReLU
\end{itemize}

\subsection{Optimization}
We train our model to maximize the likelihood of the ground truth motion token sequences via teacher forcing. We use the following training hyperparameters:
\begin{itemize}
    \item Number of training steps: 600000
    \item Batch size: 256
    \item Learning rate schedule: Linear decay
    \item Initial learning rate: 0.0006
    \item Final learning rate: 0.0 
    \item Optimizer: AdamW
    \item Weight decay: 0.6
\end{itemize}

\subsection{Inference}
We found nucleus sampling \citep{NucleusSampling}, commonly used with language models, to be helpful for improving sample quality while maintaining diversity.
Here we set the top-$p$ parameter to 0.95.

\begin{figure}
    \centering
    \includegraphics[width=0.45\linewidth]{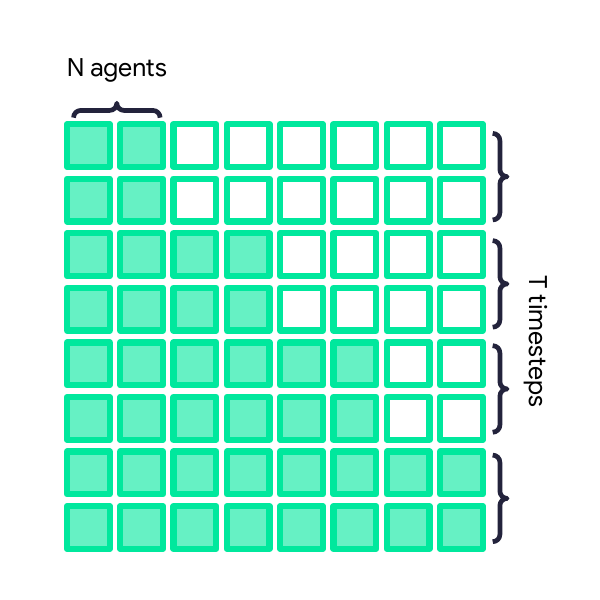}
    \caption{Masked causal attention between two agents during training. We flatten the agent and time axes, leading to an $NT \times NT$ attention mask. The agents may attend to each other's previous motion tokens (solid squares) but no future tokens (empty squares).}
    \label{fig:causal_attention}
\end{figure}

\section{Metrics descriptions}  \label{sec:metrics_desciptions}
\subsection{WOMD metrics}
All metrics for the two WOMD \citep{WOMD} benchmarks are evaluated at three time steps (3, 5, and 8 seconds) and are averaged over all object types to obtain the final value. 
For joint metrics, a scene is attributed to an object class (vehicle, pedestrian, or cyclist) according to the least common type of agent that is present in that interaction, with cyclist being the rarest object class and vehicles being the most common.
Up to 6 trajectories are produced by the models for each target agent in each scene, which are then used for metric evaluation.

\paragraph{mAP \& Soft mAP}
mAP measures precision of prediction likelihoods and is calculated by first bucketing ground truth futures of objects into eight discrete classes of intent: straight, straight-left, straight-right, left, right, left u-turn, right u-turn, and stationary.

For marginal predictions, a prediction trajectory is considered a ``miss'' if it exceeds a lateral or longitudinal error threshold at a specified timestep $T$. 
Similarly for joint predictions, a prediction is considered a ``miss'' if none of the $k$ joint predictions contains trajectories for all predicted objects within a given lateral and longitudinal error threshold, with respect to the ground truth trajectories for each agent.
Trajectory predictions classified as a miss are labeled as a false positive.
In the event of multiple predictions satisfying the miss criteria, consistent with object detection mAP metrics, only one true positive is allowed for each scene, assigned to the highest confidence prediction. All other predictions for the object are assigned a false positive.

To compute the mAP metric, bucket entries are sorted and a P/R curve is computed for each bucket, averaging precision values over various likelihood thresholds for all intent buckets results in the final mAP value.
Soft mAP differs only in the fact that additional matching predictions (other than the most likely match) are ignored instead of being assigned a false positive, and so are not penalized in the metric computation.

\paragraph{Miss rate}
Using the same definition of a ``miss’’ described above for either marginal or joint predictions, miss rate is a measure of what fraction of scenarios fail to generate \emph{any} predictions within the lateral and longitudinal error thresholds, relative to the ground truth future. 

\paragraph{minADE \& minFDE} 
minADE measures the Euclidean distance error averaged over all timesteps for the closest prediction, relative to ground truth. In contrast, minFDE considers only the distance error at the final timestep. For joint predictions, minADE and minFDE are calculated as the average value over both agents.

\subsection{Prediction overlap}  \label{sec:prediction_overlap_description}
As described in \citep{JFP}, the WOMD \citep{WOMD} overlap metric only considers overlap between predictions and ground truth.
Here we use a \emph{prediction overlap} metric to assess scene-level consistency for joint models.
Our implementation is similar to \citep{JFP}, except we follow the convention of the WOMD challenge of only requiring models to generate $(x, y)$ waypoints; headings are inferred as in \citep{WOMD}.
If the bounding boxes of two predicted agents collide at any timestep in a scene, that counts as an overlap/collision for that scene.
The final prediction overlap rate is calculated as the sum of per-scene overlaps, averaged across the dataset.

\section{Additional evaluation}

\begin{table*}\centering
\scalebox{0.94}{
\begin{tabular}{r|r|r|r|r|r|r|r|r}
\cmidrule[\heavyrulewidth]{2-9}
 &
 \multicolumn{4}{|c|}{Ensemble} &
 \multicolumn{4}{|c|}{Single Replica} \\
 \toprule
 Freq. (Hz)                   & minADE ($\downarrow$) & minFDE ($\downarrow$) & MR ($\downarrow$) & mAP ($\uparrow$) & minADE ($\downarrow$) & minFDE ($\downarrow$) & MR ($\downarrow$) & mAP ($\uparrow$) \\ \midrule 
 0.125   & 0.9120 & 2.0634 & 0.4222 & 0.2007 & 1.0681 (0.011) & 2.4783 (0.025) & 0.5112 (0.007) & 0.1558 (0.007) \\
 0.25   & 0.9083 & 2.0466 & 0.4241 & 0.1983 & 1.0630 (0.009) & 2.4510 (0.025) & 0.5094 (0.006) & 0.1551 (0.006) \\
 0.5   & 0.8931 & 2.0073 & 0.4173 & 0.2077 & 1.0512 (0.009) & 2.4263 (0.022) & 0.5039 (0.006) & 0.1588 (0.004) \\
 1   & 0.8842 & 1.9898 & 0.4117 & 0.2040 & 1.0419 (0.014) & 2.4062 (0.032) & 0.5005 (0.008) & 0.1639 (0.005) \\
 2   & \textbf{0.8831} & \textbf{1.9825} & \textbf{0.4092} & \textbf{0.2150} & \textbf{1.0345} (0.012) & \textbf{2.3886} (0.031) & \textbf{0.4943} (0.006) & \textbf{0.1687} (0.004) \\
\bottomrule
\end{tabular}
}

\caption{Joint prediction performance across varying interactive attention frequencies on the WOMD interactive validation set. Displayed are \emph{scene-level} joint evaluation metrics. For the single replica metrics, we include the standard deviation (across 8 replicas) in parentheses.}
\label{table:interact_att_ablation}
\end{table*}

\begin{table*}\centering
\scalebox{0.94}{
\begin{tabular}{r|r|r|r|r|r|r|r|r}
\cmidrule[\heavyrulewidth]{2-9}
 &
 \multicolumn{4}{|c|}{Ensemble} &
 \multicolumn{4}{|c|}{Single Replica} \\
 \toprule
 \# Rollouts                   & minADE ($\downarrow$) & minFDE ($\downarrow$) & MR ($\downarrow$) & mAP ($\uparrow$) & minADE ($\downarrow$) & minFDE ($\downarrow$) & MR ($\downarrow$) & mAP ($\uparrow$) \\ \midrule 
1 & 1.0534 & 2.3526 & 0.5370 & 0.1524 & 1.9827 (0.018) & 4.7958 (0.054) & 0.8182 (0.003) & 0.0578 (0.004) \\
2 & 0.9952 & 2.2172 & 0.4921 & 0.1721 & 1.6142 (0.011) & 3.8479 (0.032) & 0.7410 (0.003) & 0.0827 (0.004) \\
4 & 0.9449 & 2.1100 & 0.4561 & 0.1869 & 1.3655 (0.012) & 3.2060 (0.035) & 0.6671 (0.003) & 0.1083 (0.003) \\
8 & 0.9158 & 2.0495 & 0.4339 & 0.1934 & 1.2039 (0.013) & 2.7848 (0.035) & 0.5994 (0.004) & 0.1324 (0.003) \\
16 & 0.9010 & 2.0163 & 0.4196 & 0.2024 & 1.1254 (0.012) & 2.5893 (0.031) & 0.5555 (0.005) & 0.1457 (0.003) \\
32 & 0.8940 & 2.0041 & 0.4141 & 0.2065 & 1.0837 (0.013) & 2.4945 (0.035) & 0.5272 (0.005) & 0.1538 (0.004) \\
64 & 0.8881 & 1.9888 & 0.4095 & 0.2051 & 1.0585 (0.012) & 2.4411 (0.033) & 0.5114 (0.005) & 0.1585 (0.004) \\
128 & 0.8851 & 1.9893 & 0.4103 & 0.2074 & 1.0456 (0.012) & 2.4131 (0.033) & 0.5020 (0.006) & 0.1625 (0.004) \\
256 & 0.8856 & 1.9893 & \textbf{0.4078} & 0.2137 & 1.0385 (0.012) & 2.3984 (0.031) & 0.4972 (0.007) & 0.1663 (0.005) \\
512 & \textbf{0.8831} & \textbf{1.9825} & 0.4092 & \textbf{0.2150} & \textbf{1.0345} (0.012) & \textbf{2.3886} (0.031) & \textbf{0.4943} (0.006) & \textbf{0.1687} (0.004) \\
\bottomrule
\end{tabular}
}

\caption{Joint prediction performance across varying numbers of rollouts per replica on the WOMD interactive validation set. Displayed are \emph{scene-level} joint evaluation metrics. For the single replica metrics, we include the standard deviation (across 8 replicas) in parentheses.}
\label{table:num_rollouts_ablation}
\end{table*}

\paragraph{Ablations.}
\cref{table:interact_att_ablation,table:num_rollouts_ablation} display joint prediction performance across varying interactive attention frequencies and numbers of rollouts, respectively.
In addition to the ensembled model performance, single replica performance is evaluated. 
Standard deviations are computed for each metric over 8 independently trained replicas.

\paragraph{Scaling analysis.}

\cref{table:scaling_ablation} displays the performance of different model sizes on the WOMD interactive split, all trained with the same optimization hyperparameters.
We vary the number of layers, hidden size, and number of attention heads in the encoder and decoder proportionally.
Due to external constraints, in this study we only train a single replica for each parameter count.
We observe that a model with 27M parameters overfits while 300K underfits.
Both the 1M and 9M models perform decently.
In this paper, our main results use 9M-parameter replicas.

\begin{table}\centering
\scalebox{1.0}{
\begin{tabular}{r|r|r|} \toprule
 Parameter count                  & Miss Rate ($\downarrow$) & mAP ($\uparrow$) \\ \midrule 
 300K   & 0.6047 & 0.1054 \\
 1M  & 0.5037 & 0.1713 \\
 9M   & 0.4972 & 0.1663  \\
 27M   & 0.6072 & 0.1376  \\
\bottomrule
\end{tabular}
}
\caption{Joint prediction performance across varying model sizes on the WOMD interactive validation set. Displayed are \emph{scene-level} joint mAP and miss rate for 256 rollouts for a single model replica (except for 9M which displays the mean performance of 8 replicas).}
\label{table:scaling_ablation}
\end{table}

\paragraph{Latency analysis.}
\cref{table:latency} provides inference latency on the latest generation of GPUs across different numbers of rollouts.
These were measured for a single-replica joint model rolling out two agents.

\begin{table}[h]
\centering
\begin{tabular}{l|l} \toprule
Number of rollouts & Latency (ms) \\ \midrule
16          & 19.9 (0.19)        \\
32          & 27.5 (0.25)        \\
64          & 43.8 (0.26)        \\
128         & 75.8 (0.23)       \\
256         & 137.7 (0.19)      \\
\bottomrule
\end{tabular}
\caption{Inference latency on current generation of GPUs for different numbers of rollouts of the joint model. We display the mean and standard deviation (in parentheses) of the latency measurements for each setting.}
\label{table:latency}
\end{table}

\section{Visualizations}

In the supplementary zip file, we have included GIF animations of the model's greatest-probability predictions in various scenes.
Each example below displays the associated scene ID, which is also contained in the corresponding GIF filename.
We describe the examples here.

\subsection{Marginal vs. Joint}
\begin{itemize}
    \item \texttt{Scene ID: 286a65c777726df3} \\
    \textbf{Marginal:} The turning vehicle and crossing cyclist collide. \\
    \textbf{Joint:} The vehicle yields to the cyclist before turning.
    \item \texttt{Scene ID: 440bbf422d08f4c0} \\
    \textbf{Marginal:} The turning vehicle collides with the crossing vehicle in the middle of the intersection. \\
    \textbf{Joint:} The turning vehicle yields and collision is avoided.
    \item \texttt{Scene ID: 38899bce1e306fb1} \\
    \textbf{Marginal:} The lane-changing vehicle gets rear-ended by the vehicle in the adjacent lane. \\
    \textbf{Joint:} The adjacent vehicle slows down to allow the lane-changing vehicle to complete the maneuver.
    \item \texttt{Scene ID: 2ea76e74b5025ec7} \\
    \textbf{Marginal:} The cyclist crosses in front of the vehicle leading to a collision. \\ \textbf{Joint:} The cyclist waits for the vehicle to proceed before turning.
    \item \texttt{Scene ID: 55b5fe989aa4644b} \\
    \textbf{Marginal:} The cyclist lane changes in front of the adjacent vehicle, leading to collision. \\
    \textbf{Joint:} The cyclist remains in their lane for the duration of the scene, avoiding collision.
\end{itemize}

\subsection{Marginal vs. Conditional}
``Conditional'' here refers to temporally causal conditioning as described in the main text.

\begin{itemize}
    \item \texttt{Scene ID: 5ebba77f351358e2} \\
    \textbf{Marginal:} The pedestrian crosses the street as a vehicle is turning, leading to a collision. \\
    \textbf{Conditional:} When conditioning on the vehicle's turning trajectory as a query, the pedestrian is instead predicted to remain stationary.
    \item \texttt{Scene ID: d557eee96705c822} \\
    \textbf{Marginal:} The modeled vehicle collides with the lead vehicle. \\
    \textbf{Conditional:} When conditioning on the lead vehicle's query trajectory, which remains stationary for a bit, the modeled vehicle instead comes to a an appropriate stop.
    \item \texttt{Scene ID: 9410e72c551f0aec} \\
    \textbf{Marginal:} The modeled vehicle takes the turn slowly, unaware of the last turning vehicle's progress. \\
    \textbf{Conditional:} When conditioning on the query vehicle's turn progress, the modeled agent likewise makes more progress. \\
    \item \texttt{Scene ID: c204982298bda1a1} \\
    \textbf{Marginal:} The modeled vehicle proceeds slowly, unaware of the merging vehicle's progress. \\
    \textbf{Conditional:} When conditioning on the query vehicle's merge progress, the modeled agent accelerates behind.
    
\end{itemize}
    
\subsection{Temporally Causal vs. Acausal Conditioning}

\begin{itemize}
    \item \texttt{Scene ID: 4f39d4eb35a4c07c} \\
    \textbf{Joint prediction:} The two modeled vehicles maintain speed for the duration of the scene. \\
    \textbf{Conditioning on trailing agent:} \\
        - \textbf{Temporally causal:} The lead vehicle is indifferent to the query trailing vehicle decelerating to a stop, proceeding along at a constant speed. \\
        - \textbf{Acausal:} The lead vehicle is ``influenced'' by the query vehicle decelerating. It likewise comes to a stop. Intuitively, this is an incorrect direction of influence that the acausal model has learned. \\
    \textbf{Conditioning on lead agent:} \\
        - \textbf{Temporally causal:} When conditioning on the query lead vehicle decelerating to a stop, the modeled trailing vehicle is likewise predicted to stop. \\ 
        -\textbf{Acausal:} In this case, the acausal conditional prediction is similar to the temporally causal conditional. The trailing vehicle is predicted to stop behind the query lead vehicle.
\end{itemize}

\end{document}